\documentclass{article}
\usepackage{spconf,amsmath,graphicx}
\usepackage{subfigure}

\title{Who and Where: People and Location Co-Clustering}
\twoauthors
  {Zixuan Wang}
	{Stanford University\\
	zxwang@stanford.edu}
  {Jinyun Yan}
	{Rutgers, New Brunswick\\
	jinyuny@cs.rutgers.edu}

\begin{document}
\ninept
\maketitle
\begin{abstract}
%The goal of the image clustering is to group semantically related images together. %This task is very important, particularly in the era with an immense volume of user uploaded online photos.  
In this paper, we consider the clustering problem on images where each image contains patches in people and location domains.  We exploit the correlation between people and location domains, and proposed a semi-supervised co-clustering algorithm to cluster images. Our algorithm  updates the correlation links at the runtime, and produces clustering in both domains simultaneously. We conduct experiments in a manually collected dataset and a Flickr dataset. The result shows that the such correlation improves the clustering performance. 
%We propose a semi-supervised co-clustering algorithm to cluster images which takes advantage of the correlation between people and location domains. 
% While many previous work ignore the correlations between patches from different domains, we propose a co-clustering framework, which combines the visual appearance of patches in each domain and cross-domain relations. %The objective of the clustering becomes minimizing the variance within the cluster in each domain,  at the same time maximizing the consistency across both domains. 
%We design a semi-supervised kernel k-means algorithm to approximate the solution.  
%Experimental evaluations show that the convergence is reached fast and the clustering performance is promising.
\end{abstract}


\section{Introduction}
\label{sec:intro}
Given a large corpus of images, we want to cluster them such that images semantically related are grouped in one cluster. Semantics of an image refer to the information that image carries. For example, the face on the image is usually used to identify \emph{who}. The background of the image refers to the location \emph{where} the person was.  All components together can convey what story has happened.  In our case,  we focus on two entities: \emph{who} and \emph{where}.

%Online social networks host huge volumes of images nowadays.  Such explosive amount of digital photos on the web makes the clustering task crucial, in particular, when users apply image search over a person or a location. Combining with timestamps, reliable people and location clusters are also very helpful to tell stories of physical events through image corpus. 

While there has been considerable work on automatic face recognition \cite{CaoFace2010,YinAn2011} in images and even a modest effort on location recognition \cite{CityScaleLandmarkIdentification11,ChenG11}, the coupling of the two is basically unexplored.
% though people and locations are highly inter-correlated.  
An image which contains both people and location implies the co-occurrence of instances in two domains.   
 %A collection of such images can shed light on the clustering on each domain. 
For example,  multiple photos taken at the same private location increase the confidence that similar faces on those photos are from a same person.  Within a short time window, the same person on several photos indicates the affinity of locations. %In reality, considering a trip or an outdoor activity,  people and location are co-occurred frequently% on photos.   

\begin{figure*}[!ht]
  \begin{center}
    \includegraphics[width=.8\textwidth]{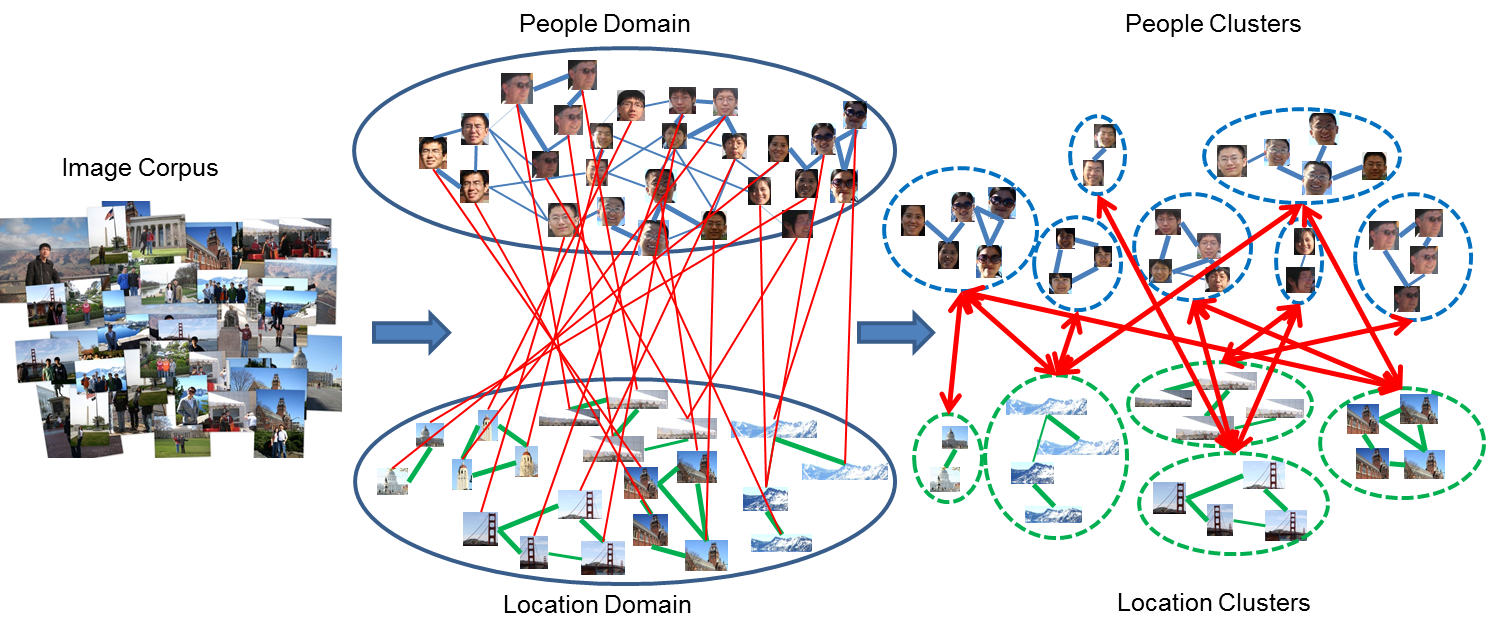}
  \end{center}
\vspace{-0.1in}
  \caption{The framework of people and locations co-clustering. The red lines represent the co-occurrence relations.}
  \label{fig:pipeline}
\vspace{-0.1in}
\end{figure*}

Our framework, shown in Fig. \ref{fig:pipeline}, consists of two domains: people and locations. We take into account  the inter-relation between two domains to enhance clustering in each domain. Three types of relations in people and location domains are considered: (1) \emph{ people-people } (2) \emph{location-location} (3) \emph{people-location}. A set of image patches is extracted and described in each domain. The similarity between patches within each domain is defined based on the visual appearances. 
%We then take into account the co-occurrence relations across \emph{ people-location} domains to enhance the clustering in each domain.  
The co-occurrence constraints are satisfied if patches from two domains appear in a same image.  This relationship reflects the consistency of clustering results which is not embodied from visual appearances in a single domain.
  
We formulate the clustering task as an optimization problem which aims to minimize the within cluster distances and maximize the consistency across domains. We show this problem can convert to the semi-supervised kernel k-means clustering similar to \cite{Kulis05}. However, we generate clustering results for two domains at the same time. During the iterative clustering process, constraints across domains and within domains keep updated. The main idea is that the clustering result in one domain can aid the clustering in the other domain. We validate our approach with photos gathered from personal albums and a set of public photos crawled from Flickr. 
%When dealing with the set of online photos, the scalability of the framework becomes a challenge. Considering search people over billion photos, it's almost impossible to store all images in one machine and never mention the case when the application requires real-time response. Therefore it is important to have a distributed algorithm. We develop a scalable solution of our framework and apply Map-Reduce on both the pre-process and alternative clustering without any loss of accuracy. We conduct the Map-Reduce framework on online datasets and observe a significant improvement with respect to the running time. 

Our contributions are threefold: 1) we propose a co-clustering algorithm for image clustering, focusing on people and locations; the algorithm couples both domains and explores underlying cross-domain relations; 2) our algorithm can simultaneously produce the clustering results of people and location, and outperforms clustering separately on each domain and the baseline co-clustering algorithm; 3) our algorithm is formulated as an optimization problem, which can be solved by through semi-supervised kernel k-means. It is robust and converges fast in practice.

%The paper is organized as follows: Section \ref{sec:rel} outlines the recent work on the people and location clustering.  Section \ref{sec:cocluster} sketches the system framework , and describes the problem definition, formulation and  the detail of algorithm. Section \ref{sec:evaluation} shows the setup of our experiments and the results. Section \ref{sec:conclusion} concludes the paper and discusses the future work.
\vspace{-0.1in}

\section{Related Work}
\label{sec:format}
\label{sec:rel}
Face is an important kind of visual objects in images, which is crucial to identify people. %It is well-structured and contains abundant information, therefore, it can provide accurate links between images. 
In recent years, there have been a lot of efforts in face detection \cite{990517}, recognition \cite{CaoFace2010,YinAn2011} and clustering \cite{Berg2004}. The basic idea is to either represent a face as one or multiple feature vectors, or parameterize the face based on some template or deformable models. In addition to treating faces as individual objects, some researchers have been seeking for help from context information, such as background, people co-occurrence, etc. Davis et al. \cite{Davis:2005:TCF:1101149.1101257} developed a context-aware face recognition system that exploits GPS-tags, timestamps, and other meta-data. %Song et al. \cite{Song2006} proposed an adaptive scheme to combine face and clothing features based on the timestamps. 
Lin et al. \cite{Lin:2010:JPE:1886063.1886083} proposed a unified framework to jointly recognize the people, location and event in a photo collection based on a probabilistic model.

Most location clustering algorithms are relying on the bag of words model \cite{Philbin07}. %First of all, interest points are detected \cite{mikolajczyk2004scale} and described by an invariant descriptor \cite{lowe2004distinctive}. Then the descriptors are  quantized into a vocabulary of visual words by approximate k-means \cite{Philbin07}. Metadata associated with images are also exploited to help the location clustering. 
Large-scale location clustering has been recently demonstrated in \cite{Li:2008:MRL:1478392.1478428,Zheng2009Tour}, which use the GPS information to reduce the large-scale task down into a set of smaller tasks. Hays et al.\ \cite{hays2008im2gps} proposed an algorithm for estimating a distribution over geographic locations from the query image using a purely data-driven scene matching approach. They leveraged a dataset of over 6 million GPS-tagged images from the Flickr. When the temporal data is available in the corpora, it also helps to localize sequences of images. %In \cite{kalogerakis2009image}, a prior of human travel patterns is added, which is learned from millions of images, to infer geographic location for sequences of time-stamped images. Chen and Grauman \cite{ChenG11} leveraged the time information and travel patterns within a Hidden Markov Model (HMM) to robustly estimate locations for sequences of tourist photos.

However, clustering in people and location domains are usually treated as separate tasks. Location patches in photos with faces are not well exploited. While GPS information of the photo is not easily accessible,  we propose a co-clustering algorithm, which simply use patches of the photo itself to discover the correlation in these two domains.

%K-means is one of the most popular clustering algorithms. A major drawback to k-means is that it can not separate clusters that are non-linearly separable in the input space. Two recent approaches have emerged for tackling such a problem. One is kernel k-means \cite{Dhillon:2004:KKS:1014052.1014118}, in which points are mapped to a higher-dimensional feature space using a nonlinear function. The other approach is the spectral clustering algorithm \cite{Ng01onspectral}, which uses the eigenvectors of an affinity matrix to obtain a clustering of the data. 
%A popular objective function used in spectral clustering is to minimize the normalized cuts \cite{Shi_2000_3808}. Kulis et al. \cite{Kulis05} showed that the spectral clustering can be formulated as a weighted kernel k-means clustering problem.
\section{Our Approach}
\label{sec:cocluster}
In this section, we present the co-clustering framework to simultaneously cluster images in people and location domains. We have two major steps. The first step is pre-processing. We extract face and location patches from the corpus of images, and compute the visual features.  The next step is co-clustering. The people-people, location-location and people-location relations are generated and updated.  We describe the detail for each step below.

\vspace*{-0.1in}
\subsection{Pre-processing}
\vspace*{-0.1in}
We here describe how to extract features from people and location domains, and discover the relations between both domains.

\emph{People Domain: }
We use Viola-Jones face detector \cite{990517} to extract face patches from an image. To obtain high accuracy, a nested detector is applied to reduce the false positive rate.  %From the collection of images, 
Every face will have a corresponding face patch.  All face patches are normalized to the same size. 
We adopt the algorithm in \cite{Uricar-Franc-Hlavac-VISAPP-2012} to detect seven facial landmarks from each extracted face patch. 
For each input face patch, four landmarks (outer eye corners and mouth corners) are registered to the pre-defined positions using the perspective transform.  Then all seven facial landmarks are aligned by the computed perspective transform. For each landmark, two SIFT descriptors of different scales are extracted to form the face descriptor. We build a face graph over all face patches in the image collection.  In the graph,  each vertex represents a face patch. The weight of the edge is the similarity of face descriptors of two face patches.

\emph{Location Domain}
For each image, Hessian affine covariant detector \cite{mikolajczyk2004scale} is used to detect interest points. The SIFT descriptor \cite{lowe2004distinctive} is extracted on every interest point. The method similar to the work of Heath et al. \cite{imageweb2010} is used to discover the shared locations in the image collection.  
The content-based image retrieval \cite{Philbin07} is applied to find top related images, and avoid quadratic pairwise comparisons. Lowe's ratio test \cite{lowe2004distinctive} is used to find the initial correspondences. RANSAC \cite{fischler1981random} is used to estimate the affine transform between a pair of images and compute feature correspondences between images.
For every location patch, two types of features are extracted: a bag of visual words \cite{Sivic03} and a color histogram. The bag of words descriptor summarizes the frequency that prototypical local SIFT patches occur. It captures the appearance of component objects. For images taken in an identical location, this descriptor will typically provide a good match.The color histogram characterizes certain scene regions well. These two types of features are concatenated to represent the location patch.
A location graph is built similarly to the face graph. Each vertex in the graph represents a location patch. The weight of the edge is the similarity of location descriptors of two location patches.

\emph{Inter-relations across Domains}
To co-cluster across the people and location domains, several basic assumptions are made as follows.

\textbf{Cannot Match Link.}  One person cannot appear twice in one image. Therefore, there is a \emph{cannot match link} between a pair of face patches in the same image. Here we do not consider the exceptions like the photo collage or mirrors in the image.  If two locations are far away according to the ground truth e.g. GPS signals,  and two face patches appear in these two locations during a short time period, there is a \emph{cannot match link}  between this pair of patches. This assumption comes from that people cannot teleport within a short time period, for example, one people cannot appear in San Francisco and in New York within an hour.

\textbf{Must Match Link.} Two location patches are connected by a \emph{must match link} if there is an affine transform found between them in the location graph construction. Because the links verified by RANSAC have high accuracy, we trust that they connect patches in the same location. Two location patches are connected by a \emph{must match link} if they appear in the same image.  Two different buildings may appear in the same image, therefore, in our setting, one location is defined as an area which may contain different backgrounds.  Two location patches are connected by a \emph{must match link} if they co-occur with the same people within a short time period. This assumption also comes from the fact that people cannot move too fast.

\textbf{Possible Match Link.}  Two face patches that appear in the same location but not in the same image probably belong to the same people, due to the strong co-occurrence between the location and the face. 
%If there are places that a person takes more than one photos, the clustering results in the location domain can help cluster the people domain, through the strong co-occurrence relations. 
This is true if the place has special meaning to the person, for example, his/her home or office, where he/she visits frequently. However, the assumption is not always true. For example, at tourist attractions, every people would take photos there. Therefore, the locations do not contribute much for the clustering in people domain. A weight is needed for the locations to distinguish the public locations and private locations. Private location is more helpful for clustering in people domain, while public location will introduce many noise.

\vspace*{-0.1in}
\subsection{Problem Formulation}
\vspace*{-0.1in}
We formulate our people and location co-clustering as an optimization problem. Given a set of feature vectors $\mathcal{X}=\{\mathbf{x}_i\}_{i=1}^n$, the goal of the standard k-means in each domain is to find a $k$-way disjoint partitioning $(S_1,\dots,S_k)$ such that the following objective is minimized:
\begin{equation}
f_{\mathrm{kmeans}}=\sum_{c=1}^k\sum_{\mathbf{x}_i\in S_c}||\mathbf{x}_i-\mathbf{m}_c||^2
\end{equation}
where $\mathbf{m}_c$ is the cluster center of $c$. 
%The objective can be rewritten using the fact that:
%\begin{equation}
%\sum_{c=1}^k\sum_{\mathbf{x}_i\in S_c}2||\mathbf{x}_i-\mathbf{m}_c||^2=\sum_{c=1}^k\sum_{\mathbf{x}_i,\mathbf{x}_j\in S_c}\frac{||\mathbf{x}_i-\mathbf{x}_j||^2}{|S_c|}
%\end{equation}
The matrix $E$ is defined as pairwise squared Euclidean distances among the data points, such that 
$E_{ij}=||\mathbf{x}_i-\mathbf{x}_j||^2$. We introduce an indicator vector $\mathbf{z}_c$ for the cluster $S_c$.
\begin{equation}
\mathbf{z}_c(i) = \left\{ \begin{array}{rl}
 1 &\mbox{ if $i\in S_c$} \\
 0 &\mbox{ if $i\not\in S_c$}
       \end{array} \right.
\end{equation}
where $\mathbf{z}_c^{\mathrm{T}}\mathbf{z}_c$ is the size of cluster $S_c$, and $\mathbf{z}^{\mathrm{T}}_cE\mathbf{z}_c$ gives the sum of $E_{ij}$ over all $\mathbf{x}_i$ and $\mathbf{x}_j$ in $S_c$. Now the matrix $\tilde{Z}$ is defined such that the $c$th column of $\tilde{Z}$ is equal to $\mathbf{z}_c/(\mathbf{z}_c^{\mathrm{T}}\mathbf{z}_c)^{1/2}$. $\tilde{Z}$ is an orthonormal matrix, $\tilde{Z}^{\mathrm{T}}\tilde{Z}=I_k$. 
%and the objective $f_{\mathrm{kmeans}}$ is:
%\begin{equation}
%\begin{aligned} 
%& \underset{\tilde{Z}}{\text{minimize}}
%& & \mathrm{tr}(\tilde{Z}^{\mathrm{T}}E\tilde{Z}) \\ 
%& \text{subject to} 
%& & \tilde{Z}^{\mathrm{T}}\tilde{Z}=I_k
%\end{aligned}
%\end{equation}
Let $N_F$ be the number of face patches and $N_L$ be the number of location patches. $k_F$ is the number of face clusters and $k_L$ is the number of location clusters. By considering the relations between people and location domains, we write the objective as: 
\begin{equation}
\label{eqn:formulation}
\begin{aligned} 
& \text{minimize}
& & f_F+f_L-f_{FL}-f_{LF} \\ 
& \text{subject to} 
& & f_F=\mathrm{tr}(\tilde{Z}_F^{\mathrm{T}}E_F\tilde{Z}_F), \quad \tilde{Z}_F^{\mathrm{T}}\tilde{Z}_F=I_{k_F},\\
&&& f_L=\mathrm{tr}(\tilde{Z}_L^{\mathrm{T}}E_L\tilde{Z}_L), \quad \tilde{Z}_L^{\mathrm{T}}\tilde{Z}_L=I_{k_L},\\
&&& f_{FL}=\sum_{i=1}^t\mathrm{tr}(M_i^{\mathrm{T}}M_i), \quad M_i=\tilde{Z}_F^{\mathrm{T}}C_{FL}^{\mathrm{T}}T_i\tilde{Z}_L, \\
&&& f_{LF}=\mathrm{tr}(N^{\mathrm{T}}N), \quad N=\tilde{Z}_F^{\mathrm{T}}C_{FL}^{\mathrm{T}}P\tilde{Z}_L.\\
\end{aligned}
\end{equation}
$E_F$ and $E_L$ are pairwise squared Euclidean distance matrices in people and location domains. To integrate the must match constraints and cannot match constraints, the distance of the must match link is set to $0$ and the distance of the cannot match link is set to $+\infty$. $f_F$ and $f_L$ with constraints $\tilde{Z}_F^{\mathrm{T}}\tilde{Z}_F=I_{k_F}$ and $\tilde{Z}_L^{\mathrm{T}}\tilde{Z}_L=I_{k_L}$ are the standard k-means optimization problems in people and location domains respectively. 

The binary people-location co-occurrence matrix $C_{FL}$ is defined as: the $i$th column of $C_{FL}$ is the location patches that co-occur with the face patch $i$. For example, if the first column of $C_{FL}$ is $(0,0,1,0,1,0,\dots)^{\mathrm{T}}$, which means the first face patch co-occurs with the third and the fifth location patches in the same image.

$C_{FL}\tilde{Z}_F$ is a clustering of location patches which is based on the face clustering result $\tilde{Z}_F$. Our goal is to maximize the consistency between the location clustering $\tilde{Z}_L$ and $C_{FL}\tilde{Z}_F$. Location patches are weighted differently to reflect different semantic meanings of the people and location interactions. It is not difficult to discover the similarity between the definitions of $f_{FL}$ and $f_{LF}$ except the weight matrix $T_i$ and $P$. $f_{FL}$ optimizes the consistency that locations co-occur with the same people during a short time period should be one location. $T_i$ is a $N_L\times N_L$ binary diagonal matrix that non-zero entries on the diagonal indicate these location patches are taken within a short time period. For example, $T_i=\mathrm{diag}(0,1,0,1,1,\dots)$ means the second, the fourth and the fifth location patches have similar timestamps. There are $t$ time constraints that are automatically learned from the meta-data of images. 

$f_{LF}$ optimizes the consistency that private locations are useful to identify people. $P$ is a $N_L\times N_L$ diagonal weight matrix. It defines a score for each location patches. The private locations have larger weights and the private locations have small weights. The diagonal matrix $P$ is defined as:
\begin{equation}
\label{eqn:p}
P_{ii}=\frac{\log (k_F/N_{F_{L_i}})}{\log (k_F)}
\end{equation}
where $P_{ii}$ approximates $0$ at public locations such as landmarks and it is approximate $1$ at private locations. $L_i$ is the location cluster that $i$ belongs to. $N_{F_{L_i}}$ is the number of people appear in location $L_i$.
% The definition of $P$ depends on the clustering results of face patches $\tilde{Z}_F$. The matrix $W_{LF}$ is defined as $W_{LF}=C_{LF}\tilde{Z}_LP\tilde{Z}_L^{\mathrm{T}}C_{LF}^{\mathrm{T}}$, which incorporate the semi-supervised information from the location graph.

\vspace*{-0.1in}
\subsection{Alternative Optimization}
\vspace*{-0.1in}
The optimization problem (\ref{eqn:formulation}) is not convex when the optimization variables involve $\tilde{Z}_F$ and $\tilde{Z}_L$. Therefore, we use the alternative optimization by fixing variables in one domain and optimize on other variables and do this iteratively. When fixing variables, e.g. $\tilde{Z}_F$. The problem becomes a semi-supervised kernel k-means problem, which can be solved easily. We solve the problem following this sequence until the convergence: $\tilde{Z}_L\rightarrow \tilde{Z}_F\rightarrow P\rightarrow \tilde{Z}_L\rightarrow \tilde{Z}_F\rightarrow P\rightarrow \cdots$. The first $\tilde{Z}_F$ and $\tilde{Z}_L$ are computed using the standard kernel k-means without cross-domain relations. After the initial clustering results are known, the weight matrix $P$ can be computed using equation (\ref{eqn:p}) and in the following iteration, the semi-supervised kernel k-means is used to integrate the cross-domains relations.
\vspace*{-0.1in}
\subsubsection{Semi-supervised Kernel K-means}
\vspace*{-0.1in}
We now briefly describe the existing semi-supervised kernel k-means algorithm \cite{Kulis05}. The objective is written as the minimization of:
\begin{equation}
\sum_{c=1}^k\sum_{\mathbf{x}_i,\mathbf{x}_j\in S_c}\frac{||\mathbf{x}_i-\mathbf{x}_j||^2}{|S_c|}-\sum_{\substack{\mathbf{x}_i,\mathbf{x}_j\in \mathcal{M}\\c_i=c_j}}\frac{2w_{ij}}{|S_c|}+\sum_{\substack{\mathbf{x}_i,\mathbf{x}_j\in \mathcal{C}\\c_i=c_j}}\frac{2w_{ij}}{|S_c|}
\end{equation}
where $\mathcal{M}$ is the set of must match link constraints, $\mathcal{C}$ is the set of cannot match link constraints, $w_{ij}$ is the penalty cost for violating a constraint between $\mathbf{x}_i$ and $\mathbf{x}_j$ , and $c_i$ refers to the cluster label of $\mathbf{x}_i$. The first term in this objective function is the standard k-means objective function, the second term is a reward function for satisfying must match link constraints, and the third term is a penalty function for violating cannot match link constraints. The penalties and rewards are normalized by cluster size: if there are two points that have a cannot match link constraint in the same cluster, we will penalize higher if the corresponding cluster is smaller. Similarly, we will reward higher if two points in a small cluster have a must match link constraint. Thus, we divide each $w_{ij}$ by the size of the cluster that the points are in.

Let $A$ be the similarity matrix $A_{ij}=\mathbf{x}_i^{\mathrm{T}}\mathbf{x}_j$ and let $\tilde{A}$ be the matrix such that $\tilde{A}_{ij}=A_{ii}+A_{jj}$. Then, $E=\tilde{A}-2A$. By replacing $E$ in the trace minimization, the problem is equivalent to the minimization of $\mathrm{tr}(\tilde{Z}^{\mathrm{T}}(\tilde{A}-2A-2W)\tilde{Z})$. We calculate $\mathrm{tr}(\tilde{Z}^{\mathrm{T}}\tilde{A}\tilde{Z})$ as $2\mathrm{tr}(A)$, which is a constant and can be ignored in the optimization. This leads to a maximization of $\mathrm{tr}(\tilde{Z}^{\mathrm{T}}(A+W)\tilde{Z})$. If we define a matrix $K=A+W$, our problem is expressed as a maximization of $\mathrm{tr}(\tilde{Z}^{\mathrm{T}}K\tilde{Z})$ and is mathematically equivalent to unweighted kernel k-means \cite{Dhillon:2004:KKS:1014052.1014118}. 
%The iterative solution of the kernel k-means in shown in Algorithm \ref{alg:kkm}.
%\begin{algorithm}                      % enter the algorithm environment
%\caption{KERNEL-KMEANS ($K$, $k$, $t_{max}$)}          % give the algorithm a caption
%\label{alg:kkm}    
%\textbf{Input:} $K$: kernel matrix, $k$: number of clusters, $t_{max}$: maximum number of iterations\\ \textbf{Output:} $\{\pi_c\}_{c=1}^k$: final partitioning of the points
%\begin{algorithmic}                    % enter the algorithmic environment.
%\STATE Initialize the $k$ clusters $\{\pi_c^{(0)}\}_{c=1}^k$ randomly
%\WHILE {not converge or $t_{max}>t$}
%\STATE For each point $\mathbf{x}_i$ and every cluster $c$, compute $d(\mathbf{x}_i,\mathbf{m}_c)=K_{ii}-\frac{2\sum_{x_j\in \pi_c}K_{ij}}{\sum_{x_j\in \pi_c}1}+\frac{\sum_{x_j,x_l\in \pi_c}K_{jl}}{(\sum_{x_j\in \pi_c}1)^2}$
%\STATE Find $c^{*}(\mathbf{x}_i)=\mathrm{argmin}_cd(\mathbf{x}_i,\mathbf{m}_c)$
%\STATE Update clusters as $\pi_c^{(t+1)}=\{\mathbf{x}_i: c^{*}(\mathbf{x}_i)=c\}$
%\ENDWHILE
%\STATE Return $\{\pi_c^{(t+1)}\}_{c=1}^k$
%\end{algorithmic}
%\end{algorithm}
\subsubsection{Alternative Optimization}

If $\tilde{Z}_F$ is fixed and $\tilde{Z}_L$ is optimized. The objective $f_{LF}$ can be written as:
\begin{equation}
f_{LF}=\mathrm{tr}(\tilde{Z}_L^{\mathrm{T}}P^{\mathrm{T}}C_{FL}\tilde{Z}_F\tilde{Z}_F^{\mathrm{T}}C_{FL}^{\mathrm{T}}P\tilde{Z}_L)
\end{equation}
The objective $f_{FL}$ can be written as:
\begin{equation}
f_{FL}=\sum_{i=1}^t\mathrm{tr}(\tilde{Z}_L^{\mathrm{T}}T_i^{\mathrm{T}}C_{FL}\tilde{Z}_F\tilde{Z}_F^{\mathrm{T}}C_{FL}^{\mathrm{T}}T_i\tilde{Z}_L)
\end{equation}
% If $\tilde{Z}_F$ is fixed and $\tilde{Z}_L$ is optimized. The objective $f_{LF}$ can be rewritten as using the fact that $\mathrm{tr}(AB)=\mathrm{tr}(BA)$.
% \begin{eqnarray}
%  f_{LF} &=& \mathrm{tr}(\tilde{Z}_F^{\mathrm{T}}PC_{LF}\tilde{Z}_L\tilde{Z}_L^{\mathrm{T}}C_{LF}^{\mathrm{T}}P^{\mathrm{T}}\tilde{Z}_F)      \nonumber \\
%    &=& \mathrm{tr}(\tilde{Z}_L^{\mathrm{T}}C_{LF}^{\mathrm{T}}P^{\mathrm{T}}\tilde{Z}_F\tilde{Z}_F^{\mathrm{T}}PC_{LF}\tilde{Z}_L)
% \end{eqnarray}
We obtain the following optimization problem:
\begin{equation}
\begin{aligned} 
& \text{maximize}
& & \mathrm{tr}(\tilde{Z}_L^{\mathrm{T}}(2A_L+\sum_{i=1}^tW_{Li}+Q_L)\tilde{Z}_L) \\ 
& \text{subject to}
& & W_{Li}=T_i^{\mathrm{T}}C_{FL}\tilde{Z}_F\tilde{Z}_F^{\mathrm{T}}C_{FL}^{\mathrm{T}}T_i, \\
&&& Q_L=P^{\mathrm{T}}C_{FL}\tilde{Z}_F\tilde{Z}_F^{\mathrm{T}}C_{FL}^{\mathrm{T}}P, \\
&&& \tilde{Z}_L^{\mathrm{T}}\tilde{Z}_L=I_{k_L}.
\end{aligned}
\end{equation}
where $A_L$ is the affinity matrix in the location domain. This optimization problem can be solved by setting the kernel matrix $K_L=2A_L+\sum_{i=1}^tW_{Li}+Q_L$. Similarly, if $\tilde{Z}_L$ is fixed and $\tilde{Z}_F$ is optimized. Using the fact that $\mathrm{tr}(AB)=\mathrm{tr}(BA)$ we can rewrite the $f_{LF}$ and $f_{FL}$, and obtain the following optimization problem:
%The objective $f_{LF}$ can be rewritten using the fact that $\mathrm{tr}(AB)=\mathrm{tr}(BA)$ as:
%\begin{equation}
%f_{LF}=\mathrm{tr}(\tilde{Z}_F^{\mathrm{T}}C_{FL}^{\mathrm{T}}P\tilde{Z}_L\tilde{Z}_L^{\mathrm{T}}P^{\mathrm{T}}C_{FL}\tilde{Z}_F)
%\end{equation}
%The objective $f_{FL}$ can be written as:
%\begin{equation}
%f_{FL}=\sum_{i=1}^t\mathrm{tr}(\tilde{Z}_F^{\mathrm{T}}C_{FL}^{\mathrm{T}}T_i\tilde{Z}_L\tilde{Z}_L^{\mathrm{T}}T_i^{\mathrm{T}}C_{FL}\tilde{Z}_F)
%\end{equation}
%We obtain the following optimization problem:
\begin{equation}
\begin{aligned} 
& \text{maximize}
& & \mathrm{tr}(\tilde{Z}_F^{\mathrm{T}}(2A_F+\sum_{i=1}^tW_{Fi}+Q_F)\tilde{Z}_F) \\ 
& \text{subject to}
& & W_{Fi}=C_{FL}^{\mathrm{T}}T_i\tilde{Z}_L\tilde{Z}_L^{\mathrm{T}}T_i^{\mathrm{T}}C_{FL}, \\
&&& Q_F=C_{FL}^{\mathrm{T}}P\tilde{Z}_L\tilde{Z}_L^{\mathrm{T}}P^{\mathrm{T}}C_{FL}, \\
&&& \tilde{Z}_F^{\mathrm{T}}\tilde{Z}_F=I_{k_F}.
\end{aligned}
\end{equation}
where $A_F$ is the affinity matrix in the face domain. This optimization problem can be solved by setting the kernel matrix $K_F=2A_F+\sum_{i=1}^tW_{Fi}+Q_F$.
\vspace{-0.1in}
\section{Evaluations}
\label{sec:evaluation}
We conduct experiments on two datasets to validate our approach. The first dataset contains images collected from personal albums with labeled ground truth. The second one uses a larger dataset crawled from online photo service: Flickr.  We choose K-means with constraints~\cite{kmeans-constraints} as the baseline algorithm.  We also compare the performance of clustering on the each single domain by normalize cut~\cite{Shi_2000_3808} and Kmeans without any constraint. We use the RandIndex~\cite{rand1971} to evaluate the performance of the clustering.
%, which is defined as follows:
%\begin{equation}
%R=\frac{a+b}{0.5n(n-1)}
%\end{equation}
%where $a$ is the number of pairs that have the same label in the ground truth and also have the same label in the clustering result; $b$ is the number of pairs that have different labels in the ground truth and have different labels 
%in the clustering result. Intuitively, $a+b$ can be considered as the number of agreements between the ground truth and the clustering result. Larger value of $R$ implies better clustering result.
%Two experiments are conducted to validate our approach. 
%The first experiment uses images collected from personal albums and the second experiment uses online images downloaded from Flickr.
\vspace*{-0.1in}
\subsection{Personal Albums}
\vspace*{-0.1in}
This dataset contains $111$ images collected from personal albums. In total it has $11$ people and $13$ locations. In the location domain, the top $50$ image candidates are selected for the pairwise geometric verification. For each image, the bounding box of matched interest points is extracted as the location patch. A bag of words histogram ($1000$ visual words), $256$-bin color histogram are extracted from each location patch. The dimension size of features in the location domain is $1,256$. We use a weight ratio of $1:1$ for BoW:color features. All feature vectors are $L2$ normalized. In total, there are $146$ face patches and $266$ location patches.

\begin{figure}[!ht]
  \begin{center}
    \subfigure[People domain]{\includegraphics[width=0.23\textwidth, height=0.2\textwidth]{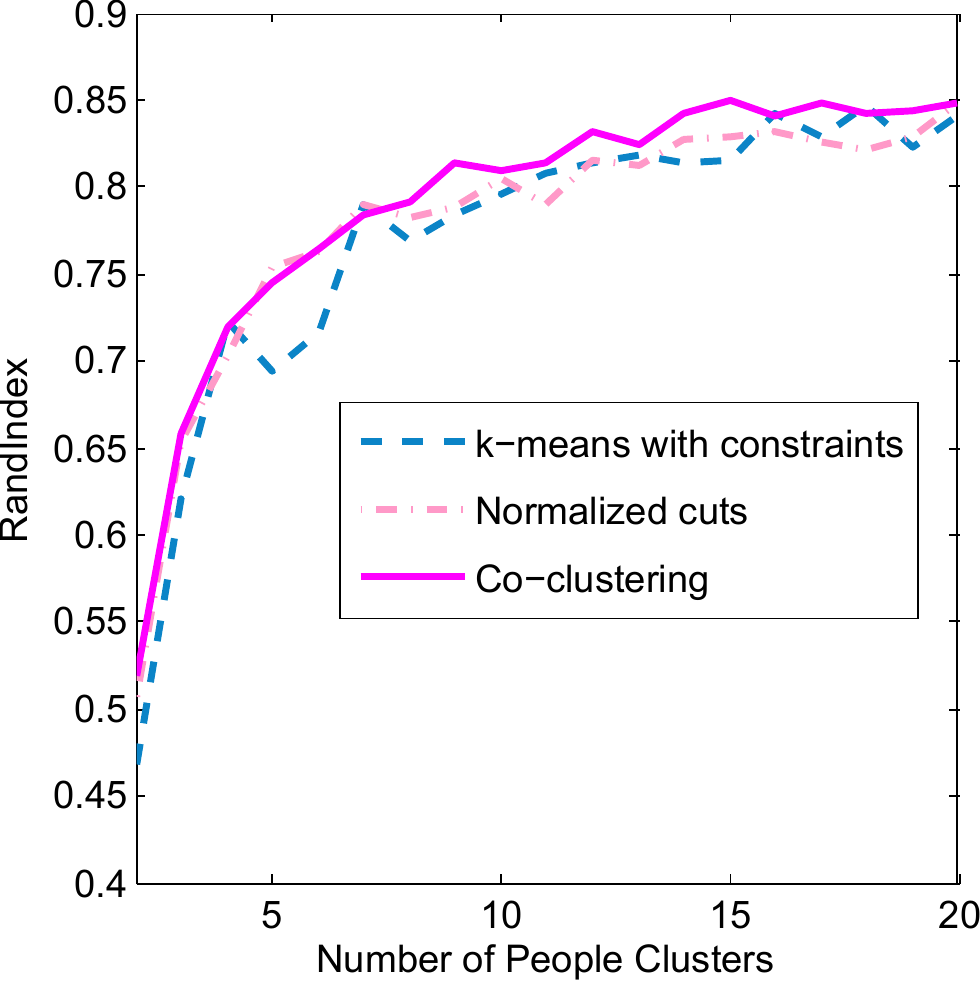}}
    \subfigure[Location domain]{\includegraphics[width=0.23\textwidth,height=0.2\textwidth]{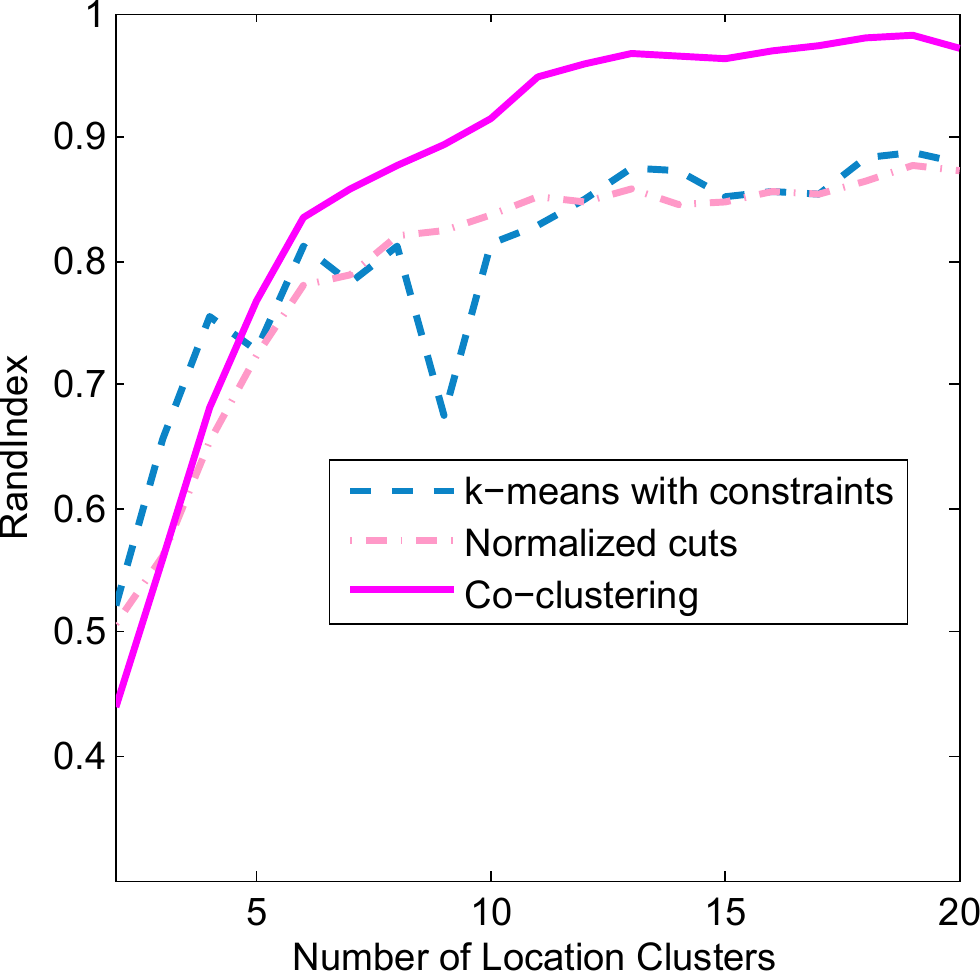}}
  \end{center}
\vspace{-0.2in}
  \caption{RandIndex on the personal dataset.}
  \label{fig:personal}
\vspace{-0.1in}
\end{figure}

In the dataset, each image associates a timestamp in the Exif header. The mean-shift \cite{1055330} is used to cluster images in the time sequence and a matrix $T_i$ is defined for each cluster of images. We cluster the face and location patches using the normalized cuts based on their appearance features as the baseline. K-means with constraints are also compared by adding the initial must match links and cannot match links in each domain. Figure \ref{fig:personal} shows results in the people domain and the location domain. %It shows that our co-clustering algorithm works the best. In particular, the performance gain is big in the location domain, and the result is more smooth in the people domain.  
  
%The convergence of our Co-clustering algorithm is reached less than ten iterations, which is quite fast. 
From Figure \ref{fig:personal}, we observe the steady improvement on the clustering results when the number of clusters is larger than 2. The k-means with constraints are quite sensitive to the number of clusters. The best RandIndex values of methods across all K values are ordered as: Co-clustering, k-means-with-constraints and Normalize cuts. The values for these methods except Co-clustering do not vary much. The performance gain of Co-clustering in the location domain is very significant. It's mainly resulted from the must match link within the location domain. For the people domain, the difference in the clustering performance is very big, however, the steady increase over K is still promising. %The improvement is caused by the help of the links between private locations and people.
% the clustering improvements in the location domain are mainly from the must links within the location domain and the clustering improvements in the people domain are mainly from the help of the private locations.

\subsection{Online Photo Sets}
\vspace*{-0.1in}
\emph{Dataset preparation}: We use 140 names of public figures to query Flickr and filter out images without geo-location information. In total, we collect 53,800 images. We then filter out images without faces. The ground truth of the people domain is obtained directly from names. The ground truth of the location is obtained by clustering the longitude and latitude associated with images. We use the agglomerative clustering to discover location clusters. We consider each geo-location data including the longitude and the latitude as a point in the two dimensional space. In this dataset, we set the number of locations to be $100$.  %Initially, we have $n$ points and assign them to $n$ different clusters. In each iteration of the clustering algorithm, we merge two clusters if the distance between two clusters is the minimum among all pairs of clusters. We keep merging clusters until the minimum distance in each iteration is above a threshold or the number of clusters we want to obtain is reached. 

\begin{figure}[!ht]
  \begin{center}
    \subfigure[People domain]{\includegraphics[width=0.23\textwidth,height=0.2\textwidth]{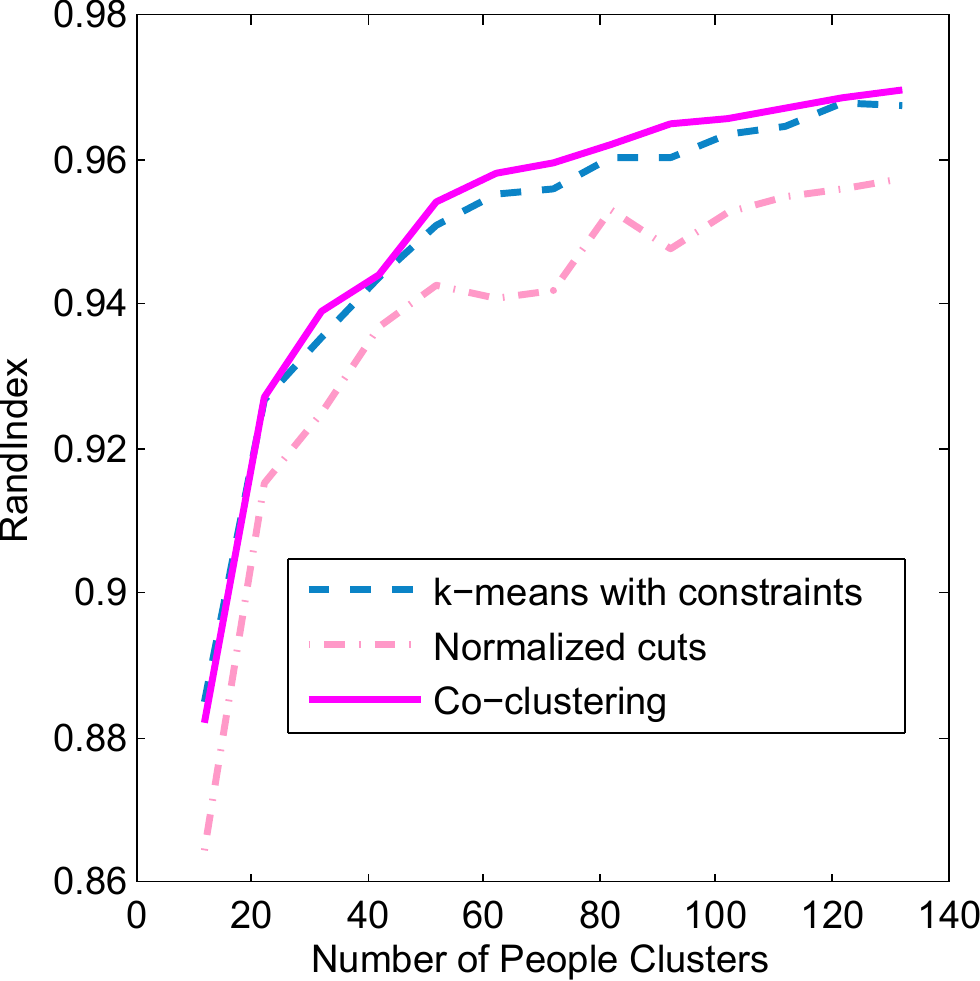}}
    \subfigure[Location domain]{\includegraphics[width=0.23\textwidth,height=0.2\textwidth]{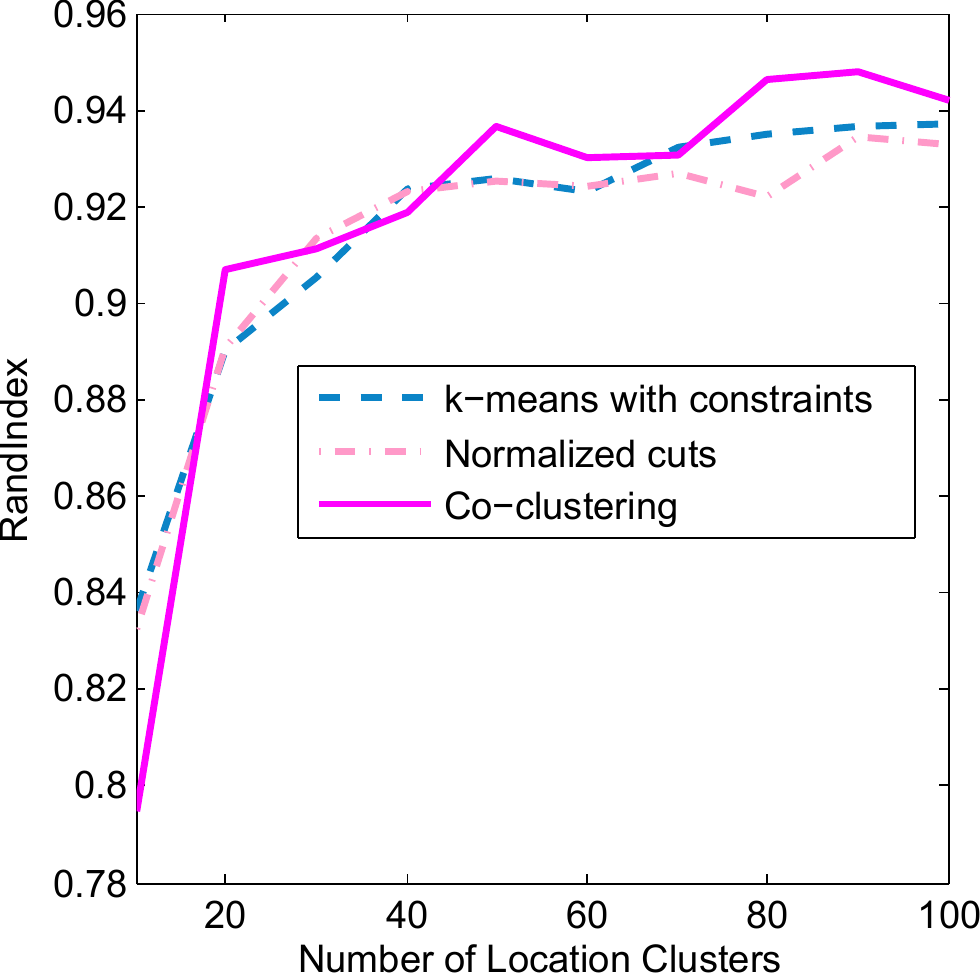}}
  \end{center}
\vspace{-0.2in}
  \caption{RandIndex on the Flickr dataset.}
  \label{fig:flickr}
\vspace{-0.1in}
\end{figure}

Figure~\ref{fig:flickr} shows RandIndex values on the people domain and the locatin domain comparing k-means, Normalized cuts and Co-clustering over different K values. The improvement is not as big as that in the personal album dataset. It is mainly caused by the noise of the image set. The ground truth of the location domain is clustered by geo-location information which is not necessary equal to the location in the image. The ground truth of the people domain could also contain noise e.g. different people with the same name may appear together within one cluster. One future work is to find efficient algorithm to deal with the noise. 

\vspace{-0.1in}

\vspace{-0.1in}
\section{Conclusion}
\vspace{-0.1in}
\label{sec:conclusion}
We present a novel algorithm to co-cluster the people and location simultaneously. The relations across domains are used to enhance the clustering in single domain. We validate our approach using two datasets, and the experiment show that our algorithm performs better than clustering in the single domain and the baseline co-clustering algorithm. 
 
%Our work gives some insight that clustering algorithms taking cross domain relations into account can achieve better results. The clustering framework in our paper can be applied into other applications that having similar cross-domain relations. In the future, we plan to handle the noise introduced by public locations, and compare our algorithm with other co-clustering algorithms. It is interesting to explore whether our algorithm works well in other areas, for example, paper and author clustering. It's very difficult to obtain the ground truth of a large scale dataset, we are seeking a good way to measure the clustering performance over very large scale photo corpus, perhaps considering the aid of crowd source. 

\bibliographystyle{abbrv}
\bibliography{sigproc}

\end{document}